\documentclass[10pt,twocolumn,letterpaper]{article}

\usepackage{iccv}              

\definecolor{iccvblue}{rgb}{0.21,0.49,0.74}
\usepackage[pagebackref,breaklinks,colorlinks,allcolors=iccvblue]{hyperref}
\usepackage{multirow}
\usepackage{bbm}
\usepackage{wrapfig}
\usepackage[accsupp]{axessibility}  
\def\paperID{8058} 
\def\confName{ICCV}
\def\confYear{2025}

\title{SynAD: Enhancing Real-World End-to-End Autonomous Driving Models through Synthetic Data Integration}

\author{
 Jongsuk Kim\textsuperscript{1}\thanks{Equal Contribution.} \quad
 Jaeyoung Lee\textsuperscript{1*} \quad
 Gyojin Han\textsuperscript{1} \quad
 Dong-Jae Lee\textsuperscript{1} \quad
 Minki Jeong\textsuperscript{2} \quad
 Junmo Kim\textsuperscript{1}
\\
 \textsuperscript{1}KAIST  \quad
 \textsuperscript{2}AI Center, Samsung Electronics
\\
 \texttt{\small \{jskpop, mcneato, hangj0820, jhtwosun, junmo.kim\}@kaist.ac.kr}
\quad 
 \texttt{\small minki6.jeong@samsung.com}
}
\begin{document}
\maketitle
\begin{abstract}
Recent advancements in deep learning and the availability of high-quality real-world driving datasets have propelled end-to-end autonomous driving. 
Despite this progress, relying solely on real-world data limits the variety of driving scenarios for training. 
Synthetic scenario generation has emerged as a promising solution to enrich the diversity of training data; however, its application within E2E AD models remains largely unexplored. 
This is primarily due to the absence of a designated ego vehicle and the associated sensor inputs, such as camera or LiDAR, typically provided in real-world scenarios.
To address this gap, we introduce SynAD, the first framework designed to enhance real-world E2E AD models using synthetic data.
Our method designates the agent with the most comprehensive driving information as the ego vehicle in a multi-agent synthetic scenario.
We further project path-level scenarios onto maps and employ a newly developed Map-to-BEV Network to derive bird’s-eye-view features without relying on sensor inputs. 
Finally, we devise a training strategy that effectively integrates these map-based synthetic data with real driving data.
Experimental results demonstrate that SynAD effectively integrates all components and notably enhances safety performance.
By bridging synthetic scenario generation and E2E AD, SynAD paves the way for more comprehensive and robust autonomous driving models.
\end{abstract}    

\section{Introduction}
\label{sec:intro} 

Autonomous vehicles are moving from research labs to roads as deep learning and comprehensive real-world driving datasets~\cite{caesar2020nuscenes} are leading to significant progress.
Recent studies leverage LiDAR~\cite{zeng2019end, khurana2022differentiable} or multi-camera images~\cite{hu2023planning,jiang2023vad,hu2022st} from these datasets to extract bird's-eye-view (BEV) features~\cite{philion2020lift,chen2022efficient,li2022bevformer} for various tasks~\cite{hu2021fiery,liu2023vision,zhang2022beverse,phan2020covernet}. 
These approaches improve the end-to-end autonomous driving (E2E AD) model's performance by incorporating perception tasks (tracking and mapping), prediction tasks (motion forecasting and occupancy prediction), and planning tasks either in parallel~\cite{weng2024drive} or in series~\cite{hu2022st, hu2023planning}, in an end-to-end manner.
To better capture the complexities of real-world driving environments, some studies~\cite{tong2023scene,zheng2025occworld} have proposed methods incorporating 3D occupancy understanding.

\begin{figure}[t]
\centering
    \includegraphics[width=\linewidth]{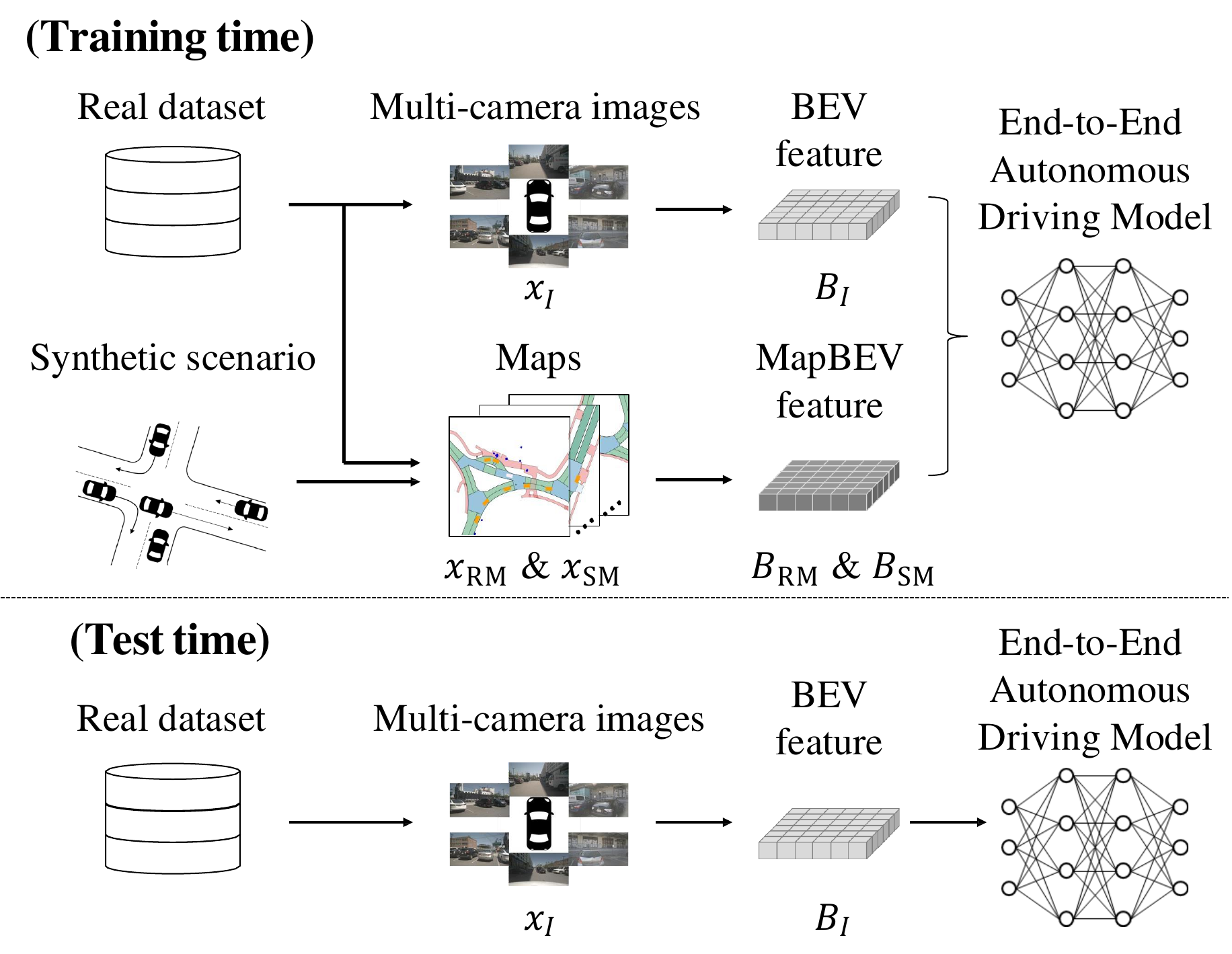}
    \caption{Conceptual illustration of SynAD. During training, both real and synthetic data are used to generate BEV and MapBEV features for the E2E AD model, while only real data is used during testing to ensure practical applicability.}
    \label{fig:intro} 
    
\end{figure}

However, relying solely on real-world datasets introduces a fundamental limitation: the high costs of data collection and labeling lead to a lack of diversity, restricting the range of scenarios available for training.
To mitigate this issue, some studies~\cite{hanselmann2022king, zhang2024chatscene} generate paths under specific conditions and utilize CARLA simulator~\cite{dosovitskiy2017carla} to acquire corresponding camera data for additional model training. 
These approaches enable robust driving in extreme situations but still have limitations since they can only operate in virtual environments.
In parallel, several studies have developed methods to generate realistic driving scenarios from real-world datasets without relying on simulators. 
These works employ logic-based~\cite{zhong2023guided}, language-guided~\cite{zhong2023language, ruan2024traffic}, and retrieval-based~\cite{ding2023realgen} methods to produce high-quality and diverse scenarios that satisfy specific conditions. 

Despite the high generative capabilities of these methods, synthetic scenarios have not been effectively integrated into real-world E2E AD model training.
A key limitation is that current scenario generation approaches yield only path-level outputs and overlook the designation of an ego vehicle. They also fail to generate the corresponding sensor inputs, such as multi-camera images and LiDAR data, which are required to establish the ego-centric perspective seen in real-world scenarios.
Consequently, this absence restricts their integration into real-world E2E AD training pipelines.

To address these challenges, we propose SynAD, a novel framework that enhances real-world E2E AD models by integrating synthetic data. SynAD comprises three key components:
First, we introduce an ego-centric scenario generation method specifically tailored for E2E AD training. 
During scenario generation, we set effective guides while designating the agent with the richest driving information as the ego vehicle. The path of the selected ego vehicle is then set as the target path and serves as additional training data for the E2E AD model.
Second, we propose a Map-to-BEV Network to integrate synthetic scenarios into the E2E AD training pipeline.
The Map-to-BEV Network encodes BEV features from maps that contain vehicle information from the synthetic scenarios, enabling this integration without relying on sensor data inputs.
Finally, we reduce the domain gap between map-based synthetic data and real driving data by also projecting real scenarios onto a map, ensuring consistent integration as shown in Figure~\ref{fig:intro}.
Moreover, by selectively utilizing features extracted from each type of map at the most suitable stage, we avoid performance degradation from integrating map data and ensure the model achieves high test time performance with image-only inputs.
Extensive ablation studies verify that each component of SynAD contributes effectively to the application of synthetic scenarios in E2E AD training.
Our main contributions are summarized as follows:
\begin{itemize}
    \item To overcome the lack of necessary sensor data and the absence of an ego-centric perspective in synthetic scenarios, we propose SynAD, a novel method that integrates synthetic data into real-world E2E AD models.
    \item SynAD introduces three key contributions: (1) ego-centric scenario generation method that transforms path-level scenarios into ego-centric maps by designating the most informative agent as the ego vehicle, (2) a Map-to-BEV Network that produces BEV features without relying on any sensor inputs, and (3) a training strategy that effectively utilizes both synthetic and real data.
    \item Extensive experiments demonstrate that SynAD outperforms existing methods, with ablation studies confirming the effectiveness of each component.
\end{itemize}
\section{Related Works}
\subsection{Traffic Scenario Generation} 
Traffic scenario generation is crucial for testing and improving autonomous driving systems by enabling safe and comprehensive validation in simulated environments. Recent studies~\cite{hanselmann2022king, xu2023diffscene, zhang2024chatscene, ding2020learning, wang2021advsim, rempe2022generating} have focused on safety-critical scenarios, which are difficult to capture in real-world driving due to cost and safety constraints. 
KING~\cite{hanselmann2022king} uses a kinematic bicycle model to derive gradients of safety-critical objectives, updating paths that make the ego vehicle more likely to cause accidents. It also improves the robustness of E2E AD in synthetic driving environments based on simulators by fine-tuning these generated scenarios. 
Beyond purely safety-critical contexts, research on controllable scenario generation~\cite{zhong2023guided, zhong2023language, jiang2023motiondiffuser, pronovost2023scenario, lu2024scenecontrol, xu2023diffscene} is also receiving great attention.
These works introduce diffusion models that allow users to specify trajectory properties (e.g., reaching a goal, following speed limits) while preserving physical feasibility and natural behaviors.
In addition, some studies~\cite{zhong2023language, ruan2024traffic, xia2024language, liu2024controllable} leverage large language models to convert user queries into realistic traffic scenarios.
RealGen~\cite{ding2023realgen} highlights a limitation in generative approaches that they often struggle to produce novel scenarios and propose combining behaviors from multiple retrieved examples for creating new scenarios.
However, employing these generated scenarios to improve real-world E2E AD models remains largely unexplored.

\begin{figure*}[t]
\centering
    \includegraphics[width=1.0\linewidth]{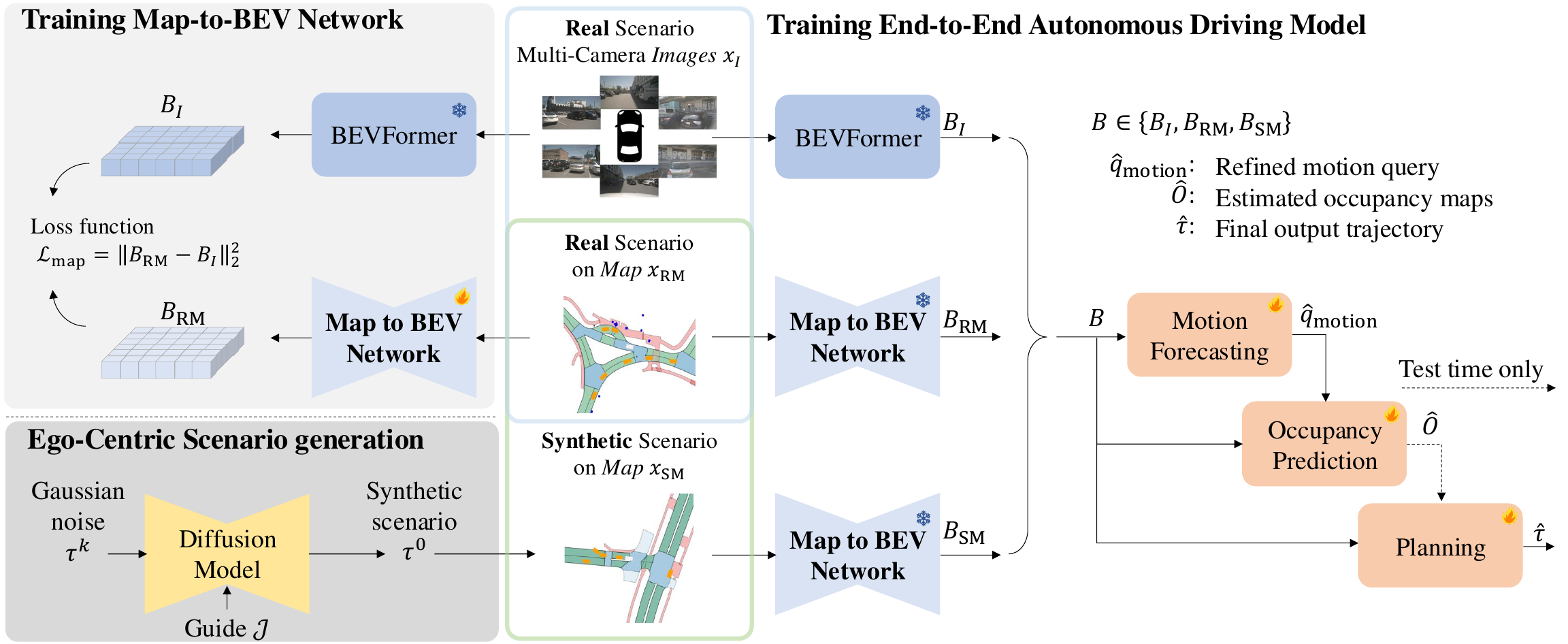}
    \caption{Overview of SynAD. We generate synthetic multi-agent scenarios and convert them into ego-centric map representations $x_\text{SM}$, while real scenarios are similarly projected as $x_\text{RM}$. 
    To train Map-to-BEV Network, we use paired data from $x_\text{RM}$ and $x_I$, ensuring that Map-to-BEV Network produces BEV feature consistent with the output of pretrained BEVFormer applied to multi-camera images. 
    The synthetic scenario $x_\text{SM}$ can be converted into BEV feature $B_\text{SM}$ without any multi-camera images using our novel Map-to-BEV network.
    In the final E2E AD framework, we selectively apply BEV features only to modules that benefit most, thereby improving overall performance.}
    \label{fig:method} 
\end{figure*}
\subsection{End-to-End Autonomous Driving}
E2E AD, particularly vision-centric approaches, has become an active area of recent research. 
Unlike conventional AD methods~\cite{chen2021data,gonzalez2015review,kendall2019learning,xu2021autonomous}, which separate perception tasks and planning, vision-centric E2E methods integrate these components into a single unified model. 
These approaches provide interpretability and safety benefits while improving performance in each downstream task through end-to-end optimization.
Planning-oriented modular design principles have driven several recent advances in E2E~AD.
ST-P3~\cite{hu2022st} trains semantic occupancy prediction and planning in an end-to-end manner. 
UniAD~\cite{hu2023planning} proposes a planning-oriented unification of tracking, online mapping, motion forecasting, occupancy prediction, and planning. 
Paradrive~\cite{weng2024drive} achieves parallel processing across these modules, boosting runtime speed by nearly threefold. 
VAD~\cite{jiang2023vad} replaces dense rasterized scene representations with fully vectorized data to boost efficiency. 
Meanwhile, OccNet~\cite{tong2023scene} and OccWorld~\cite{zheng2025occworld} explore 3D occupancy representation by segmenting the scene into structured cells with semantic labels.
Despite differences in network design and framework implementation, these methods all rely on BEV features derived from multi-camera inputs.

\section{Method}
Our method aims to enhance the E2E AD model by leveraging synthetic data.
First, we generate multi-agent driving scenarios that satisfy specific conditions and convert them into ego-centric scenarios by designating an ego vehicle and cropping a map centered around it. 
We denote this synthetic ego-centric map representation as $x_\text{SM}$, which is used in E2E AD training.
In parallel, we train the Map-to-BEV Network using $x_{\text{RM}}$, constructed by projecting real-world scenarios onto a corresponding map representation.
The Map-to-BEV Network aligns BEV features extracted from $x_{\text{RM}}$ with multi-camera images $x_I$, enabling the use of synthetic scenarios without requiring sensor inputs.
Finally, we propose a training strategy that incorporates synthetic scenarios into the E2E AD training process.
Figure~\ref{fig:method} provides an overview of our method.

\subsection{Ego-centric Scenario Generation}
\paragraph{Realistic Scenario Generation.} 
In an autonomous driving system, the trajectory of a vehicle is represented by its state $s$ at each time step $t$.
This state vector comprises four elements: position in 2D coordinates ($x,y$), speed $v$, and heading angle $\theta$, represented as $s = (x, y, v, \theta)$.
To generate realistic scenarios in ego-centric autonomous driving environments that meet desired conditions, we employ conditional diffusion models that aim to generate trajectory $\tau$, which includes the state of $M$ agents over $T$ timestamps:
\begin{equation}
\tau =  [\tau_1, \tau_2, \dots, \tau_M], 
\text{ where } \tau_i = [s_i^1, s_i^2, \dots, s_i^T]^\top,
\end{equation}
$s_i^t$ denotes the state of agent $i$ at time $t$, and $\tau \in \mathbb{R}^{T\times M\times 4}$.
The diffusion model adds Gaussian noise in a forward process and then reconstructs it in a reverse process. 
Defining $\tau^k$ as the trajectory at the $k$-th diffusion step, the forward process is defined as: 
\begin{align}
    q\left(\tau^{1:K}\ |\ \tau^0\right)&=\prod_{k=1}^{K}{q\left(\tau^k\mid\tau^{k-1}\right)}, \\
    q\left(\tau^k\ |\ \tau^{k-1}\right)&=\mathcal{N}\left(\tau^k;\sqrt{1-\beta_k}\tau^{k-1},\beta_kI\right),
\end{align}
and $\beta_k$ is the variance schedule controlling the amount of noise added at each diffusion step. 
Note that $\tau^0$ represents the clean trajectory, and $\tau^K$ represents the trajectory corrupted by random noise after $K$ diffusion steps.

To incorporate contextual information into the reverse diffusion process, we construct a composite feature $\mathbf{f}$ by aggregating the image features from the past $h$ timestamps. These image features are extracted from maps that display only the road layout and environmental context, without any vehicle depictions.
Then, the reverse diffusion process $p_\varphi$ can be represented as follows:
\begin{align}
    p_\varphi\left(\tau^{0:K}\mid\mathbf{f}\right) &= p\left(\tau^K\right)\prod_{k=1}^{K}{p_\varphi\left(\tau^{k-1}\mid\tau^k,\mathbf{f}\right)}, \\
    p_\varphi\left(\tau^{k-1}\mid\tau^k,\mathbf{f}\right) &= \mathcal{N}\left(\tau^{k-1};\mu_\varphi\left(\tau^k,k,\mathbf{f}\right), \Sigma_\varphi\left(\tau^k,k,\mathbf{f}\right)\right),
\end{align}
where $\tau^K \sim \mathcal{N}(\mathbf{0}, \mathbf{I})$  starts as random noise and is progressively denoised over $K$ steps by $p_\varphi$.
\begin{figure}[t]
    \centering
    \includegraphics[width=\linewidth]{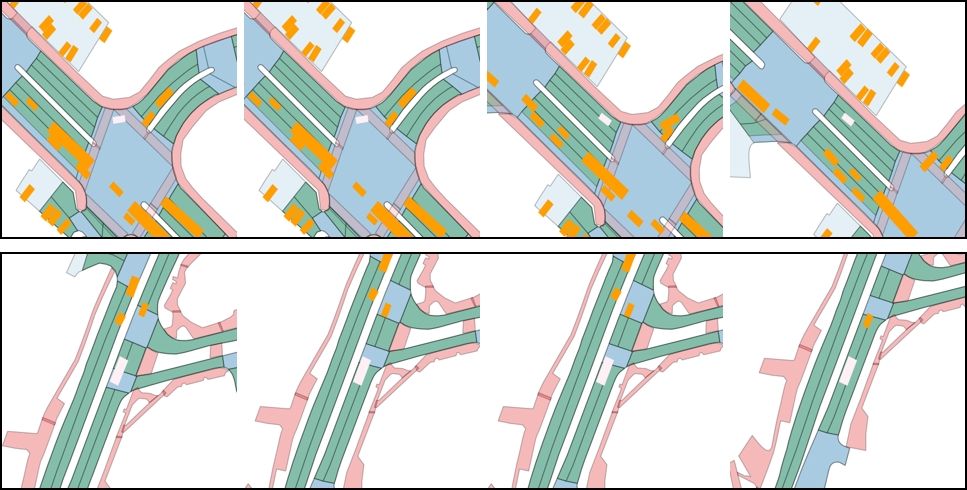}
    \caption{Examples of $x_\text{SM}$ over time. White box indicates the ego vehicle, while orange boxes denote other vehicles. The synthetic scenarios are conditioned on the existing map representation, then projected using vehicle states and size information.}
    \label{fig:exp}
\end{figure}

\paragraph{Guided Sampling.} 
To generate realistic scenarios under diverse conditions, we apply guided sampling during inference.
We define a guide $\mathcal{J} = \sum{w_{i}R_{i}}$ as the weighted sum of functions $R_{i}$ that measure rule satisfaction for the $i$-th objective. In our work, we employ three specific objectives: preventing agent collisions, preventing map collisions, and adhering to speed limits (detailed in Supp.~\ref{sup:guide}). 
We then modify the denoising process by applying the gradient of this guide as follows:
\begin{equation}\label{eq:guide}
    p_\varphi\left(\tau^{k-1}\mid\tau^k,\mathbf{f}\right) = \mathcal{N}\left(\tau^{k-1};\mu_\varphi + \nabla{\mathcal{J}(\mu_\varphi)}, \Sigma_\varphi\right).
\end{equation}
By modifying the reverse diffusion process, we can dynamically generate trajectories that satisfy each objective.

\paragraph{Ego-centric Scenario.}
\label{method:ego}
To train the autonomous driving model using multi-agent scenarios generated by the diffusion model, we first determine the ego-vehicle among the agents. 
Since synthetic scenarios should contain comprehensive driving information, we establish an ego selection rule that designates the vehicle traveling the longest distance as the ego vehicle.
Accordingly, the ego index $e$ is determined as:
\begin{equation}\label{eq:ego}
    e = \arg\max_i\sum_{t=1}^{T-1} d(s_i^{t}, s_i^{t+1}),
\end{equation}
where $d(\cdot, \cdot)$ represents the distance between positions of states.
We then crop a fixed-size area centered on the ego vehicle to form the input map $x_{\text{SM}}^t$ for timestamp $t$.

Training the planning module requires the future trajectory of the ego vehicle and the bounding boxes of other vehicles to ensure collision-free predictions. 
Since the generated trajectories are in the absolute coordinate system, we transform them into coordinate systems relative to the selected ego vehicles to align them with the real data.
The transformation aligns the driving direction of the ego vehicle with the positive $y$-axis and sets its center as the origin.
To transform an arbitrary position $(x,y)$ in the absolute coordinate system relative to the state of ego vehicle $s$, we define the transformation function as:
\begin{align}\label{eq:rot}
        T(x,y;s) = \begin{pmatrix}
                      \sin{s_\theta} & -\cos{s_\theta} \\
                      \cos{s_\theta} & \sin{s_\theta}
                    \end{pmatrix} \times 
                      \begin{pmatrix}
                          x-s_x\\
                          y-s_y
                      \end{pmatrix}.
\end{align}
The derivation of this specific form of the rotation matrix is included in the Supp.~\ref{sup:ego}.
To obtain the target path over the next $T_p$ timestamps in the ego-centric coordinate frame at time $t$, we apply the transformation $T(\cdot;s_e^t)$ to the positions of the ego vehicle. 
We also transform the heading angle relative to the ego vehicle's orientation at time $t$ as:
\begin{align}
    \mathcal{T}^t&=\left\{T\left(s_x, s_y ; s_e^t\right) \mid s= s_e^{t+t^{\prime}}, t^{\prime} \in [T_p]\right\} ,\label{eq:set_t}\\
    \Theta^t&= \left\{s_\theta - s_{e,\theta}^{t} \mid  s= s_e^{t+t^{\prime}}, t^{\prime} \in [T_p] \right\}  ,
\end{align}
where $[T_p]$ denotes the set of integers from $1$ to $T_p$. 
To process the bounding box information, we first obtain the bounding box coordinates $b_i^{t+t'}$, for each vehicle $i$ at time $t+t'$.
We then apply the transformation across other vehicles and timestamps, resulting in:
\begin{equation}\label{eq:set_b}
\mathcal{B}^t = \left\{ T(x, y;\ s_e^t) \mid (x, y) \in b_i^{t+t'}, i \in [M] \backslash \{e\}, t' \in [T_p] \right\}.
\end{equation}
Finally, each scenario provides $(\mathcal{T}^t, \Theta^t, \mathcal{B}^t, w_e, h_e)$ for the input $x_{\text{SM}}^t$ where $t \in [T-T_p]$. 
Unlike real driving datasets, the ego vehicle in synthetic data can vary in size across scenarios as shown in Figure~\ref{fig:exp}. Therefore, ego vehicle's width and height $(w_e, h_e)$ are also included in each instance. 

\subsection{Map-to-BEV Network} 
To address the absence of sensor inputs (e.g., multi-camera images or LiDAR) in synthetic scenarios, we introduce a Map-to-BEV Network $f_{B}$ that generates BEV features directly from ego-centric map inputs.
Consistent with our synthetic data generation pipeline, we derive the map input $x_{\text{RM}}$ from the real scenario.
A map encoder $f_{M}$ processes $x_{\text{RM}}$ into a spatial feature, which then serves as the key and value in a Transformer encoder. A learnable query $Q_B$ is used as the query input, producing the mapBEV feature. The encoding process is as follows:
\begin{align}
    B_{\text{RM}} &= f_{B}(Q_B, x_{\text{RM}}) \\
                  &= \operatorname{TransformerEncoder}(Q_B, f_M(x_{\text{RM}})).
\end{align}
This design enables the Transformer encoder to capture spatial relationships within the map feature, producing accurate map-based BEV representation.
Concurrently, we utilize the pre-trained BEVFormer~\cite{li2022bevformer} to extract BEV features $B_I$ from multi-camera images $x_I$, which correspond to the map input $x_{\text{RM}}$.
To align the BEV features extracted from the map ($B_\text{RM}$) with those extracted from multi-camera images ($B_I$), we employ an L2 loss function. Formally, the loss function for training the Map-to-BEV Network can be expressed as follows:
\begin{equation} 
    \mathcal{L}_{\text{map}} = \lVert B_{\text{RM}} - B_{I} \rVert_2^2. 
\end{equation}
This step allows the Map-to-BEV Network to generate BEV features without depending on sensor inputs.
Consequently, we can encode the BEV features from $x_{\text{SM}}$, enabling E2E AD model training using synthetic scenarios.
\begin{figure}[t]
    \centering
    \includegraphics[width=\linewidth]{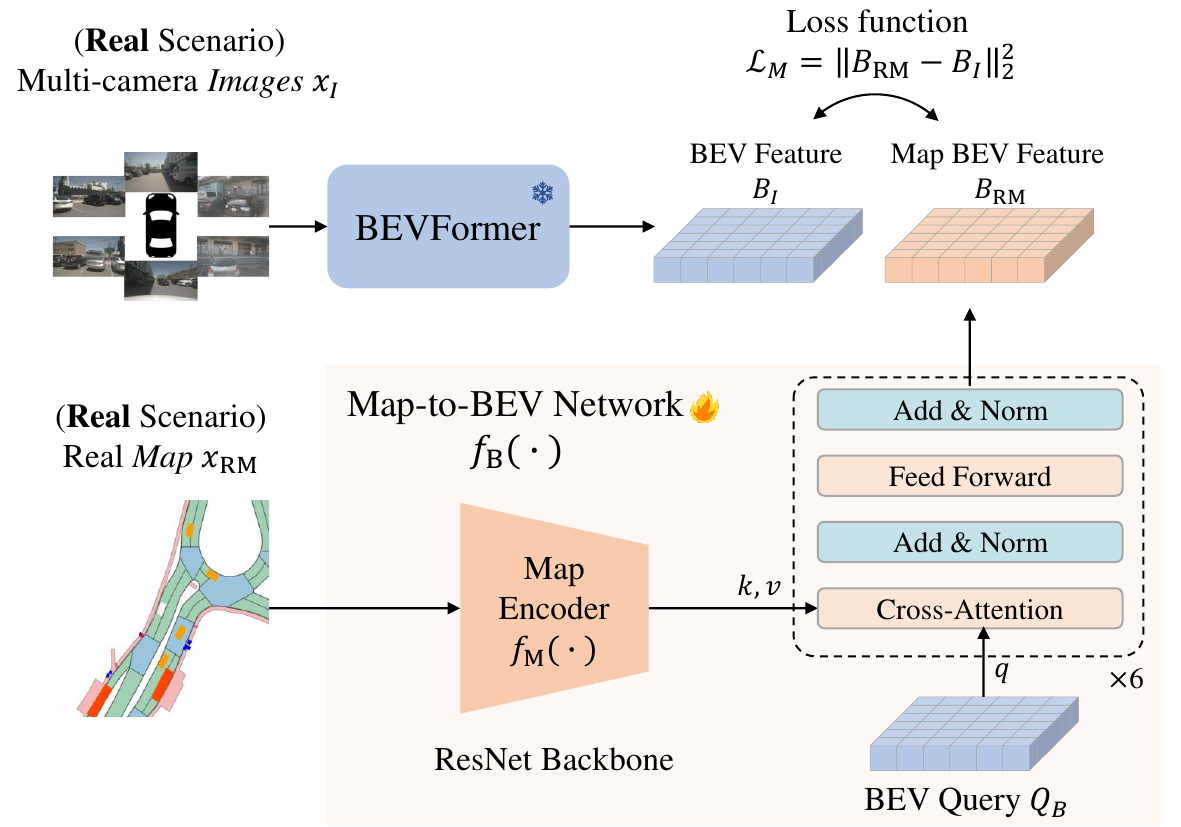}
    \caption{Overview of the Map-to-BEV training. We freeze pre-trained BEVFormer and align $B_\text{RM}$ with $B_I$, enabling the network to generate BEV representations without sensor inputs.}
    \label{fig:sup_map}
\end{figure}

\subsection{Training E2E AD with Generated Scenario}\label{method:stage2}
To train our model, we use generated data $x_{\text{SM}}$ alongside multi-camera image data $x_I$ and real data $x_{\text{RM}}$. 
The image data $x_I$ is fed into BEVFormer to produce BEV features $B_I$, while map data $x_{\text{RM}}$ and $x_{\text{SM}}$ pass through the Map-to-BEV Network to obtain $B_{\text{RM}}$ and $B_{\text{SM}}$, respectively.
E2E AD models typically comprise three main BEV-based modules: a perception module that handles tracking and mapping, a prediction module for motion forecasting and occupancy prediction, and a planning module. 
Since our map input already includes much of the perception-level information, we do not incorporate it into the perception module, and focus on integrating map data into prediction and planning modules.
In the following, we provide a brief overview of each module, with additional details in Supp.~\ref{sup:arch}.

\paragraph{Motion Forecasting.}
To predict trajectories for multi-agents over multi-timestamps with possible $N$ series of waypoints, we employ MotionEncoder and MotionDecoder based on deformable cross-attention~\cite{zhu2020deformable}. 
MotionEncoder produces a motion query embedding $q_\text{motion}$ that represents general motion patterns, agent-centered motion offsets, and relationships to ego vehicle. The MotionDecoder refines this embedding using the BEV feature, yielding multiple predicted trajectories $\hat{\mathbf{x}}_n$ and their probabilities $p_n$.
To handle multi-hypothesis output, we find trajectory $\hat{\mathbf{x}}_{n^*}$, closest to ground-truth trajectory $\mathbf{x}_\text{GT}$ by minimizing average displacement error over time. We then calculate joint negative log-likelihood loss as:
\begin{align} 
    \mathcal{L}_\text{JNLL} &= - \log \left( p_{n^*} \cdot P(\mathbf{x}_\text{GT} \mid \hat{\mathbf{x}}_{n^*}) \right), \\
    \text{where} \quad n^* &= \arg\min_n \left( \frac{1}{T} \sum_{t=1}^T \lVert \hat{\mathbf{x}}_n^{t} - \mathbf{x}_\text{GT}^{t} \rVert_2^2 \right).
\end{align} 
With the minimum final displacement error (minFDE), total motion forecasting loss $\mathcal{L}_\text{motion}$ is defined as:
\begin{align} 
    \mathcal{L}_\text{minFDE} &= \min_n \left( \lVert \hat{\mathbf{x}}_n^{T} - \mathbf{x}_\text{GT}^{T} \rVert_2^2 \right), \\
    \mathcal{L}_\text{motion} &= \lambda_\text{JNLL} \mathcal{L}_\text{JNLL} \nonumber + \lambda_\text{minFDE} \mathcal{L}_\text{minFDE},
\end{align} 
where $\lambda_\text{JNLL}$ and $\lambda_\text{minFDE}$ are corresponding loss weights.

\paragraph{Occupancy Prediction.}
An occupancy prediction module forecasts future occupancy maps.
Using embedding $\hat{q}_\text{motion}$ from motion forecasting module, we first derive temporal queries $q_\text{temp}$. 
Then, temporal queries refine a downscaled BEV feature $B_{\text{state}}^{t-1}$ through a transformer-based OccDecoder, producing updated BEV feature $B_{\text{state}}^{t}$.
Once all timestamps have been processed, we combine final BEV features with instance queries to produce occupancy maps $\hat{O}^t$.
The predicted occupancy maps $\hat{O}=\{\hat{O}^1,...,\hat{O}^T\}$ are compared with ground-truth occupancy maps $O_\text{GT}$ to compute occupancy prediction loss $\mathcal{L}_\text{occ}$, which consists of dice loss $\mathcal{L}_\text{dice}$ and binary cross-entropy loss $\mathcal{L}_\text{bce}$ as follows: 
\begin{equation} 
\mathcal{L}_{\text{occ}} = \lambda_{\text{dice}} \mathcal{L}_{\text{dice}}(\hat{O}, O_{\text{GT}}) + \lambda_{\text{bce}} \mathcal{L}_{\text{bce}}(\hat{O}, O_{\text{GT}}), 
\end{equation} 
where $\lambda_{\text{dice}}$ and $\lambda_{\text{bce}}$ are the corresponding loss weights.

\paragraph{Planning.}
Following prior works~\cite{hu2022st, hu2023planning}, we concatenate a high-level command embedding with a learnable parameter and pass them through a linear layer to form the initial planning query $q_\text{plan}$.
A Transformer-based PlanDecoder refines this query using an adapted BEV feature $B_a$ as the key and value inputs:
\begin{align}
    \hat{q}_\text{plan} &=\operatorname{PlanDecoder}(q_\text{plan}, B_a).
\end{align}
Then $\hat{q}_\text{plan}$ passes through an MLP to obtain displacement vectors $\Delta \hat{\tau} = \operatorname{MLP}(\hat{q}_\text{plan})$.
Taking the cumulative sum of these displacements across timesteps produces the final predicted trajectory $\hat{\tau}$. 
The planning loss is defined as the sum of the imitation loss and the collision loss, both computed from $\hat{\tau}$.
The imitation loss measures the L2 distance between the $\hat{\tau}$ and the ground-truth trajectory $\tau$.
For the collision loss, we obtain the ego vehicle’s bounding box at timestamp $t$ as:
\begin{equation}
    \hat{b}^t(\delta) = \operatorname{box}(\hat{\tau}^t, w_e+\delta, h_e+\delta).
\end{equation}
where $\delta$ is a safety margin.
We then compute collision loss using IoU between $\hat{b}^t(\delta)$ and each other vehicle's bounding box $b_i^t$ across all timesteps. Combining these losses for multiple values of $\delta$ yields the planning loss as below:
\begin{align}
    \mathcal{L}_{\text{col}}(\delta) &= \sum_{i,t} \operatorname{IoU}\left(\hat{b}^t(\delta), b_i^t\right), \\
    \mathcal{L}_{\text{plan}} &= {\lVert \tau - \hat{\tau} \rVert}_2^2 + \sum_{(\lambda_\delta, \delta)}\lambda_\delta \mathcal{L}_{\text{col}}(\delta).
\end{align}
Note that in training with generated scenarios, $(w_e,h_e)$ may vary, the ground-truth trajectory $\tau$ is taken from $\mathcal{T}$ (Eq.~\ref{eq:set_t}), and each bounding box $b_i^t$ is an element of $\mathcal{B}$ (Eq.~\ref{eq:set_b}).

Finally, the loss function for E2E AD training can be expressed by incorporating scaling factors as follows:
\begin{equation}\label{eq:t_loss}
    \mathcal{L}_{\text{E2E}} = \lambda_{\text{motion}}\mathcal{L}_{\text{motion}} + \lambda_{\text{occ}}\mathcal{L}_{\text{occ}} + \lambda_{\text{plan}}\mathcal{L}_{\text{plan}}.
\end{equation}

\paragraph{Map-Data Integration.}
We incorporate map-based BEV features into the motion forecasting and planning modules, where additional contextual information (e.g., road geometry, traffic structure) proves beneficial.
In contrast, occupancy prediction requires high spatial precision, making 2D map data less helpful~\cite{zheng2025occworld}; we thus exclude map inputs for this module to prevent performance degradation. 
Experimental results confirm that this selective integration avoids degrading overall performance and maintains strong test-time accuracy with image-only data.

\section{Experiments}
\subsection{Implementation Details}
\paragraph{Scenario Generation.} 
We conduct all experiments using nuScenes~\cite{caesar2020nuscenes}, a real-world driving dataset containing 1,000 scenes. Each nuScenes scene consists of 40 video frames, capturing a 20-second video at 2Hz. We set the future prediction timestamp \(T_p\) to 6, which yields 34 training instances per scene. 
For our main results, we train the model from scratch for 5 epochs while adding 500 synthetic scenes, which is equivalent to 7.5 epochs if training solely on the original nuScenes dataset. Despite this additional data, our total training cost remains lower than that of other E2E AD methods. 
For ablation studies, unless otherwise noted, we use 100 synthetic scenes for training.
To maintain sufficient interaction complexity, we exclude any instance that contains only a single driving agent.

\paragraph{Training Details.}
For training the Map-to-BEV Network, we freeze the pre-trained BEVFormer~\cite{li2022bevformer} and update only the Map-to-BEV network parameters over $20$ epochs. 
For the E2E AD model, we train each module from scratch while keeping the BEVFormer and the Map-to-BEV network frozen. 
At test time, we apply the occupancy-based optimization from UniAD~\cite{hu2023planning}.
All experiments are conducted on 8 NVIDIA RTX 4090 GPUs with batch size $1$ per GPU.
More details can be found in the Supp.~\ref{sup:detail}.

\subsection{Main Results}
We evaluate our method on the nuScenes validation set, adopting the CTG++~\cite{zhong2023language} metrics for scenario generation and the VAD~\cite{jiang2023vad} evaluation protocol for the E2E AD task, ensuring consistency with existing methods.
Details on the reporting rules can be found in Supp.~\ref{sup:other}

\paragraph{Scenario Generation.}
\begin{table}[t]
    \centering
    \resizebox{\linewidth}{!}{
        \begin{tabular}{c|ccc|ccc}
            \toprule
            \multirow{2}{*}{Method} & \multicolumn{3}{c|}{no collision $\downarrow$} & \multicolumn{3}{c}{no offroad $\downarrow$}  \\ 
            \cline{2-7}
             & rule & real & rel real & rule & real & rel real  \\
            \hline
            BITS~\cite{xu2023bits}        & 0.065 & 0.099 & 0.352 & 0.018 & 0.099 & 0.355 \\
            BITS+opt~\cite{xu2023bits}    & 0.041 & 0.070 & 0.353 & 0.005 & 0.100 & 0.358 \\
            CTG~\cite{zhong2023guided}         & 0.052 & 0.044 & 0.346 & \textbf{0.002} & 0.042 & 0.346\\
            CTG++~\cite{zhong2023language}       & 0.036 & \textbf{0.040} & 0.332 & 0.004 & \textbf{0.038} & 0.328 \\
            SynAD(Ours) & \textbf{0.033} & 0.045 & \textbf{0.330} & \textbf{0.002} & 0.040 & \textbf{0.324} \\
            \bottomrule
        \end{tabular}
    }
    \caption{Evaluation of synthetic scenarios with varying guidance.}
    \label{exp:main1}
\end{table}
In Table~\ref{exp:main1}, we evaluate our generated paths using three metrics: \textit{rule}, \textit{real}, and \textit{rel real}. The \textit{rule} metric indicates how strictly the generated trajectories adhere to given rules. The \textit{real} metric measures absolute similarity to real-world data using the Wasserstein distance, while \textit{rel real} assesses the realism of scene-level interactions between vehicles.
Our method demonstrates robust compliance with traffic constraints, as indicated by its substantial \textit{rule} score. Although it has a slightly lower \textit{real} score, suggesting a looser correspondence to exact real-world trajectories, it achieves a higher \textit{rel real} score that highlights more sophisticated multi-agent interactions.
These results show that the generated trajectories deviate from real-world paths while still capturing diverse driving behaviors, which is advantageous for building more robust E2E AD systems.

\begin{table}[t]
    \centering
    \resizebox{\linewidth}{!}{
        \begin{tabular}{l|cccc|cccc}
            \toprule
            \multirow{2}{*}{Method} & \multicolumn{4}{c|}{L2($m$) $\downarrow$} & \multicolumn{4}{c}{Colllsion Rate($\%$) $\downarrow$}\\ 
            & $1s$ & $2s$ & $3s$   & Avg.   & $1s$ & $2s$ & $3s$   & Avg.    \\
            \midrule
            ST-P3$^\dagger$~\cite{hu2022st}   & 1.33 & 2.11 & 2.90 & 2.11 & 0.23 & 0.62 & 1.27 & 0.71\\
            UniAD~\cite{hu2023planning}& 0.48 & 0.74 & 1.07 & 0.76  & 0.12 & 0.13 & 0.28 & 0.17 \\
            VAD~\cite{jiang2023vad}   & 0.41 & 0.70 & 1.05& 0.72 & 0.07 & 0.17 & 0.41 & 0.22\\
            OCCNet$^\dagger$~\cite{tong2023scene}  & 1.29 & 2.31 & 2.99 & 2.14 & 0.21 & 0.59 & 1.37 & 0.72\\
            Paradrive~\cite{weng2024drive}  & \textbf{0.25} & \textbf{0.46} & \textbf{0.74} & \textbf{0.48} & 0.14 & 0.23 & 0.39 & 0.25\\
            OCCWorld~\cite{zheng2025occworld}  & 0.32 & 0.61 & 0.98 & 0.64 & 0.06 & 0.21 & 0.47 & 0.24\\
            SynAD (Ours)  &0.52 & 0.78 & 1.10 & 0.80 & \textbf{0.04} & \textbf{0.10} & \textbf{0.20} & \textbf{0.11} \\
            \bottomrule
        \end{tabular}
    }
    \caption{Planning performance on the nuScenes validation set. $^\dagger$~denotes results evaluated under the ST-P3 metric.}
    \label{exp:main2}
\end{table}
\paragraph{Planning.} 
Table~\ref{exp:main2} presents our planning performance from two perspectives: trajectory accuracy, measured by the L2 distance error from the ground truth path, and safety, represented by the collision rate with other vehicles. 
While our SynAD model exhibits slightly higher L2 distance errors due to the broader distribution of generated behaviors, it achieves the lowest collision rate among all baselines, indicating superior collision avoidance. 
This trade-off stems from emphasizing more diverse, realistic interactions during scenario generation, which yields safer but not necessarily GT-matching trajectories. 
Since real-world driving prioritizes collision avoidance over precise path replication, our approach is particularly well-suited for practical deployment. 
Additionally, SynAD is the only method to incorporate variations in vehicle sizes during training, further enhancing its adaptability to real-world driving conditions.

\paragraph{Prediction.}
\begin{table}[t]
    \centering
    \resizebox{\linewidth}{!}{
        \begin{tabular}{l|ccc|cc}
            \toprule
            \multirow{2}{*}{Method} & \multicolumn{3}{c|}{Motion Forecasting $\downarrow$} & \multicolumn{2}{c}{Occupancy. $\uparrow$} \\ 
            & minADE & minFDE & MR & IoU-n   & IoU-f   \\
            \hline
            UniAD~\cite{hu2023planning}       & 0.75 & 1.10 & 0.166 & \textbf{61.9} & \textbf{39.7} \\
            VAD$^*$~\cite{jiang2023vad}         & 0.78 & 1.11 & 0.169 & - & - \\
            Paradrive~\cite{weng2024drive}   & 0.73 & 1.08 & 0.162 & 60.0 & 36.4  \\
            SynAD (Ours)        & \textbf{0.69} & \textbf{1.01} & \textbf{0.154} & 60.5 & 39.6 \\
            \bottomrule
        \end{tabular}
    }
    \caption{Prediction performance on the nuScenes validation set. Results reproduced in our environments. $^*$VAD does not have an occupancy prediction module.}
    \label{exp:main3}
\end{table}
Motion forecasting and occupancy prediction results provide insights into the E2E AD model's ability to interpret and anticipate the behavior of surrounding objects and agents.
The results in Table~\ref{exp:main3} show that SynAD excels in accurately predicting the movements of surrounding agents and maintains a solid understanding of environmental occupancy. 
Even when synthetic data is introduced as a new input type, the model demonstrates robust performance during testing with image-only input, validating the effectiveness of the integration strategy.

\subsection{Ablation Studies}\label{sec:abl}
\paragraph{Training Strategy.}
\begin{table}[t]
    \centering
    \resizebox{\linewidth}{!}{
        \begin{tabular}{ccc|c|ccc|cc|cc}
            \toprule
            \multicolumn{4}{c|}{Updated Modules} & \multicolumn{3}{c|}{\multirow{2}{*}{Motion Forecasting $\downarrow$}} & \multicolumn{2}{c|}{\multirow{2}{*}{Occupancy. $\uparrow$}}   & \multicolumn{2}{c}{\multirow{2}{*}{Plan.(avg.) $\downarrow$}} \\ 
            \cline{1-4}
            \multicolumn{3}{c|}{$x_\text{RM}$} & $x_\text{SM}$ & & & & & & \\ 
            \hline
            Mot. & Occ. & Plan & Plan & minADE & minFDE & MR & IoU-n & IoU-f & L2 & Col.  \\
            \hline
            & & & & 0.76 & 1.11 & 0.162 & 60.1 & 38.9 & 1.15 & 0.25 \\
            & & & \checkmark & 0.75 & 1.13 & 0.166 & 59.3 & 38.3 & 0.82 & 0.19 \\
            & & \checkmark & \checkmark & 0.77 & 1.15 & 0.168 & 60.2 & 39.0 & 0.79 & 0.18 \\
            & \checkmark & \checkmark & \checkmark & 0.77 & 1.14 & 0.167 & 58.4 & 37.6 & 0.78 & 0.18 \\
            \checkmark & & \checkmark & \checkmark & \textbf{0.73} & \textbf{1.06} & \textbf{0.157} & \textbf{60.2} & \textbf{39.2} & \textbf{0.77} & \textbf{0.14} \\
            \bottomrule
        \end{tabular}
    }
    \caption{Prediction and planning performance variations based on the incorporation of $x_\text{RM}$ and $x_\text{SM}$ in each E2E AD module.}
    \label{abl:strategy}
\end{table}
To effectively leverage the synthetic scenarios in our E2E AD framework, we use $x_\text{RM}$, the real scenario projected onto the map, as a training bridge. 
Table~\ref{abl:strategy} presents the results of this approach.
First, incorporating $x_\text{SM}$ into the planning module training significantly improves planning performance, while adding $x_\text{RM}$ provides a modest additional gain. 
However, when we extend real map to the occupancy prediction module, performance declines, suggesting that 2D map representations alone are insufficient for this task.
These results indicate that BEV features extracted from map data suffice for motion forecasting and planning but fall short for occupancy prediction. 
The latter often requires richer spatial information, as evidenced by OCCNet~\cite{tong2023scene} and OCCWorld~\cite{zheng2025occworld}, which leverage 3D data to improve performance. 
Consequently, our main training strategy updates only the motion forecasting and planning modules through the map data.

\paragraph{Scale of Synthetic Scenarios.}
\begin{table}[t]
    \centering
    \resizebox{\linewidth}{!}{
        \begin{tabular}{c|ccc|cc|cc}
            \toprule
            \multirow{2}{*}{\# Synthetic scenes} & \multicolumn{3}{c|}{Motion Forecasting $\downarrow$} & \multicolumn{2}{c|}{Occupancy. $\uparrow$}   & \multicolumn{2}{c}{Plan.(avg.) $\downarrow$} \\ 
            \cline{2-8}
             & minADE & minFDE & MR & IoU-n & IoU-f & L2 & Col.  \\
            \hline
            \multicolumn{4}{l}{\textit{Baseline}} \\
            \hline
            0   & 0.76 & 1.11 & 0.162 & 60.1 & 38.9 & 1.15 & 0.25 \\
            \hline
            \multicolumn{4}{l}{\textit{Same step (Fair comparison)}} \\
            \hline
            100 & 0.73 & 1.06 & 0.157 & \textbf{60.2} & \textbf{39.2} & \textbf{0.77} & 0.14 \\
            300 & \textbf{0.72} & \textbf{1.02} & \textbf{0.153} & 59.4 & 38.6 & 0.81 & \textbf{0.13} \\
            500 & 0.73 & 1.03 & 0.155 & 59.4 & 38.7 & 0.85 & 0.14 \\
            \hline
            \multicolumn{4}{l}{\textit{Same epoch (Longer training)}} \\
            \hline
            100 & 0.72 & 1.04 & 0.156 & 60.3 & 39.1 & \textbf{0.76} & 0.13 \\
            300 & 0.71 & 1.02 & 0.155 & 60.3 & 39.4 & 0.77 & 0.12 \\
            500 & \textbf{0.69} & \textbf{1.01} & \textbf{0.154} & \textbf{60.5} & \textbf{39.6} & 0.80 & \textbf{0.11} \\
            \bottomrule
        \end{tabular}
    }
    \caption{Performance under different numbers of generated scenes, comparing two training protocols.}
    \label{abl:scenes}
\end{table}
Table~\ref{abl:scenes} illustrates how varying their number influences performance under two training protocols: one with the same number of training steps and another with the same number of epochs. 
When no synthetic scenes are used, the model relies solely on multi-camera image data, forming our baseline. 
In the same-step protocol, incorporating a moderate amount of synthetic scene improves results, although further increases yield diminishing results.
Under the same-epoch protocol, introducing more synthetic scenes consistently enhances performance, demonstrating that the model benefits from broader coverage given sufficient training iterations.
Across both protocols, L2 distance tends to increase with more scenes, reflecting the broader distribution of synthetic scenarios. 
In particular, faster convergence under the same training steps as the baseline underscores the advantages of using the synthetic scenario.

\begin{figure*}[!t]
\centering
  \includegraphics[width=\linewidth]{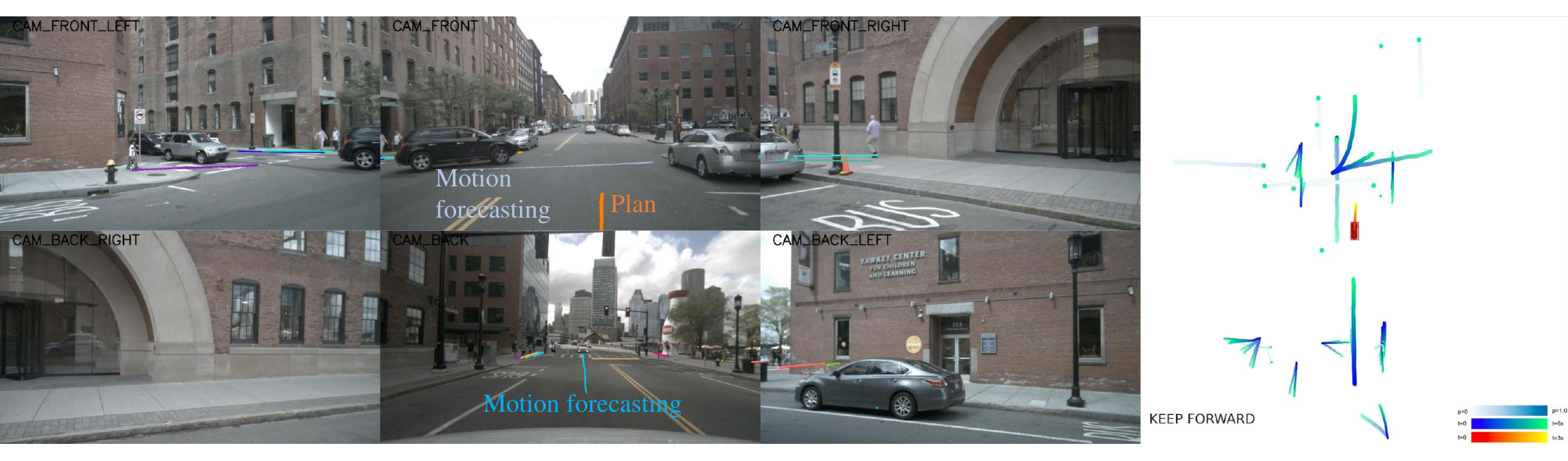}
\caption{Qualitative result of SynAD. The performance of SynAD in an urban driving scenario is presented through six views capturing the surroundings. The front and back vehicles' motion forecasting are visualized with color-coded trajectories, where warmer colors (red) indicate more immediate movements and cooler colors (blue) represent later positions.}
\label{fig:qual} 
\end{figure*}

\paragraph{Map-to-BEV Network Architecture.}
\begin{table}[t]
    \centering
    \resizebox{\linewidth}{!}{
        \begin{tabular}{c|c|c|ccc|cc}
            \toprule
            \multirow{2}{*}{Arch.} & input & $\mathcal{L}_{\text{map}}^{\text{val}}\downarrow$  & \multicolumn{3}{c|}{Motion Forecasting $\downarrow$} & \multicolumn{2}{c}{Plan.(avg.) $\downarrow$}\\ 
            & res. & $(\times 10^{-2})$ & minADE & minFDE & MR & L2 & Col.    \\
            \midrule
            SwinUNETR & 800 & 9.55 & 0.75 & 1.11 & 0.158 & 1.08 & 0.26   \\
            Ours      & 224 & \textbf{8.96} & \textbf{0.73} & \textbf{1.06} & \textbf{0.157} & \textbf{0.77} & \textbf{0.14}  \\
            \bottomrule
        \end{tabular}
    }
    \caption{Performance variations based on Map-to-BEV network architectures.}
    \label{abl:arch}
\end{table}

Table~\ref{abl:arch} shows the ablation results on the different network architectures for the Map-to-BEV Network.
We compare our model with SwinUNETR~\cite{hatamizadeh2021swin}, which preserves spatial correspondence between the map and BEV features.
One observation is that SwinUNETR requires high-resolution inputs to achieve sufficient performance, leading to higher computational costs.
After training each Map-to-BEV Network variant, we evaluate BEV feature quality using an L2 loss on the validation dataset (i.e. $\mathcal{L}_{\text{map}}^{\text{val}}$).
Then, we compare the performance of motion forecasting and planning in E2E AD training, which are influenced by map data. 
The results demonstrate that our design outperforms SwinUNETR while incurring lower computational costs.

\paragraph{Guide Composition for Scenario Generation.}
\begin{table}[t]
    \centering
    \resizebox{\linewidth}{!}{
        \begin{tabular}{ccc|cccc|cccc}
            \toprule
            \multicolumn{3}{c|}{Guide} & \multicolumn{4}{c|}{L2($m$) $\downarrow$} & \multicolumn{4}{c}{Collision Rate($\%$) $\downarrow$}\\ 
            agent & map & speed  & $1s$ & $2s$ & $3s$   & Avg.   & $1s$ & $2s$ & $3s$   & Avg.    \\
            \midrule
            \checkmark & & &  \textbf{0.48} & \textbf{0.73} & \textbf{1.05} & \textbf{0.75} & 0.05 & 0.15 & 0.35 & 0.18 \\
            \checkmark & \checkmark & &  0.49 & 0.74 & 1.06 & 0.76 & 0.05 & 0.12 & 0.27 & 0.15\\
            \checkmark & \checkmark & \checkmark &  0.50 & 0.75 & 1.07 & 0.77 & \textbf{0.05} & \textbf{0.11} & \textbf{0.26} & \textbf{0.14}\\
            \bottomrule
        \end{tabular}
    }
    \caption{Planning performance with varying guide functions for sampling synthetic scenarios.}
    \label{abl:gen}
\end{table}
Table~\ref{abl:gen} presents the impact of different guide compositions from Equation~\ref{eq:guide} on planning performance. 
The agent guide aims to prevent collisions between agents, and the map guide prevents collisions with map components, and the speed guide enforces both minimum and maximum speeds.
Incorporating the map and speed guides progressively decreases collision rates, demonstrating that these additional constraints enhance safety.
While we focus on specific guide functions, extending the approach, such as using LLM- or retrieval-based guidance, is left for future work.

\paragraph{Ego Selection Rule.}
For synthetic scenario generations, we experiment with three ego selection rules: random, dynamic, and longest.
Each rule selects vehicles with a minimum movement of $1m$ over the generated timestamps to ensure meaningful training. 
The random rule selects any vehicle meeting this criterion, while the dynamic rule designates the vehicle with the largest lateral ($x$-axis) movement. 
Lastly, the longest rule selects the ego vehicle that traveled the longest distance, following Equation~\ref{eq:ego}.
Table~\ref{abl:ego} presents both the planning performance and the average distance traveled by the ego vehicle over future timestamp $T_p$. 
When selecting the ego vehicle based on the longest rule, the model shows the lowest collision rate. Furthermore, the longest rule achieves the lowest L2 errors with sufficient trajectory distance.
Based on the results, the longest rule is set as the ego selection rule for the synthetic scenario that we incorporate into our E2E AD training.
\begin{table}[t]
    \centering
    \resizebox{\linewidth}{!}{
        \begin{tabular}{c|c|cccc|cccc}
            \toprule
            \multirow{2}{*}{Rule} & \multirow{2}{*}{Dist.} & \multicolumn{4}{c|}{L2($m$) $\downarrow$} & \multicolumn{4}{c}{Colllsion Rate($\%$) $\downarrow$}\\ 
            & & $1s$ & $2s$ & $3s$   & Avg.   & $1s$ & $2s$ & $3s$   & Avg.    \\
            \midrule
            Random   & 3.67 &0.51 & 0.77 & 1.09 & 0.79 & 0.06 & 0.11 & 0.31 & 0.16\\
            Dynamic  & 3.68 & 0.50 & 0.77 & 1.11 & 0.79 & \textbf{0.04} & \textbf{0.10} & 0.31 & 0.15\\
            Longest  & 3.70 & \textbf{0.50} & \textbf{0.75} & \textbf{1.07} & \textbf{0.77} & 0.05 & 0.11 & \textbf{0.26} & \textbf{0.14}\\
            \bottomrule
        \end{tabular}
    }
    \caption{Planning performance with different ego vehicle selection rules in synthetic scenarios. \textit{Dist.} represents the average distance traveled between consecutive frames.}
    \label{abl:ego}
\end{table}

\section{Conclusion}
We propose SynAD, a novel method that integrates synthetic scenarios into real-world E2E AD models. SynAD overcomes the limitations that previously confined such integrations to virtual environments like simulators. By utilizing map-based BEV feature encoding, we enable the training of synthetic scenarios without relying on sensor data such as multi-camera images or LiDAR data.
Also, we propose ego-centric scenario generation methods and strategic integration approaches.
Meanwhile, integrating synthetic scenarios holds significant potential for incorporation into existing E2E AD pipelines. 
We leave applying our integration strategy to other E2E AD methods for future work.
\section*{Acknowledgments}
This work was supported by Samsung Electronics Co., Ltd (IO231005-07280-01).

{
    \small
    \bibliographystyle{ieeenat_fullname}
    \bibliography{main}
}
\clearpage

\setcounter{page}{1}
\maketitlesupplementary
\setcounter{table}{0}
\renewcommand{\thetable}{\Alph{table}}

\setcounter{figure}{0}
\renewcommand{\thefigure}{\Alph{figure}}

\setcounter{equation}{0}
\renewcommand{\theequation}{\Alph{equation}}

\setcounter{section}{0}
\renewcommand{\thesection}{\Alph{section}}

\section{Ego-centric Scenario Generation}
\subsection{Diffusion Training.}
For the backbone model, we leverage a U-Net backbone with custom temporal adaptations for handling sequential features and multi-agent scenarios.
In the reverse process, the condition $\mathbf{f}$ is constructed by concatenating the map features from the past $h=1$ timestamp. These map features are encoded using a ResNet model.
We train the diffusion model on the nuScenes training set for 200 epochs using the Adam optimizer with a learning rate of $2\times10^{-4}$ and employ cosine learning rate decay.

\subsection{Guide Functions}\label{sup:guide}
\paragraph{Agent Collision.}
To prevent agents from colliding with each other, we introduce an agent collision guide function that penalizes trajectories where agents come too close.
Each agent $i$ is approximated as a circle centered at its predicted position, with a radius $r_i$ defined as half the diagonal length.
To maintain a safe separation between agents, we define a safety distance as the sum of their radius and a buffer distance $\delta$:
\begin{equation}
    d_{\text{safe},ij} =r_i+r_j+\delta, \quad \text{where} \quad \delta = 1.
\end{equation}
For each pair of agents $i$ and $j$, we denote the Euclidean distance between the two agents as $d_{ij}^t$.
Then, we define the agent collision guide for agent $i$ at timestamp $t$ as:
\begin{align}
    R_{\text{agent}, i}^t = \sum_{j\neq i} \max \left(1-\frac{d_{ij}^t}{d_{\text{safe},ij}}, 0 \right)
\end{align}
To emphasize avoiding collisions earlier in the trajectory, we apply an exponential decay weighting to the collision penalties computed across all timestamps $T$:
\begin{align}
    &R_{\text{agent},i} = \displaystyle\sum_{t=1}^T w(t) R_{\text{agent},i}^t, \\
    &\text{where} \quad w(t)=\frac{\gamma^t}{\sum_{k=1}^T \gamma^k}, \quad \gamma = 0.9. 
\end{align}
In practice, the collision guide is computed only for agents with non-zero velocity, $\mathbbm{1}_\text{agent}(i) = 1$.
This ensures that stationary agents are excluded from the collision penalty, and the final agent collision guidance is calculated overall $M$ agents as:
\begin{equation}
    R_{\text{agent}} = \sum_{i=1}^M \mathbbm{1}_\text{agent}(i) R_{\text{agent},i}.
\end{equation}

\paragraph{Map Collision.}
We introduce a map collision guide function to ensure that agents remain in drivable areas and avoid off-road regions. The function provides gradients that guide the agent back onto the road by considering the spatial relationship between off-road and on-road points within the agent's bounding box.
For each agent $i$ at timestamp $t$, we sample a set of points $P_i^t$ arranged in a $10 \times 10$ grid along the width and height within its bounding box. This set of points is then divided into an on-road set $O_i^t$ and an off-road set $F_i^t$ with as follows:
\begin{align}
    O_i^t = \{p \in P_i^t \mid \mathcal{M}(p)=1\}, \\
    F_i^t = \{p \in P_i^t \mid \mathcal{M}(p)=0\},
\end{align}
where $\mathcal{M}(p)$ returns $1$ if the point $p$ is on-road and $0$ otherwise.
The map collision guide encourages off-road points $p_{\text{off}}$ to align more closely with the on-road region by minimizing their distance to the nearest on-road point $p_{\text{on}}$. Additionally, we apply the same exponential decay weighting function $w(t)$ as defined in the agent collision guide below:
\begin{equation}
    R_{\text{map},i}^t=\sum_{p_{\text{off}} \in F_i^t}\left(1-\displaystyle\min_{p_{\text{on}}\in O_i^t}\lVert p_{\text{on}} - p_{\text{off}}\rVert_2 \right),
\end{equation}
\begin{equation}
    R_{\text{map},i} = \displaystyle\sum_{t=1}^T w(t) R_{\text{map},i}^t,
\end{equation}
$\mathbbm{1}_\text{map}(i)$ ensures the map loss is applied only to agents that are moving and partially on- and off-road(i.e., $O_i^t \neq \emptyset$ and $F_i^t \neq \emptyset$). We calculate the map collision guide as follows:
\begin{equation}
    R_{\text{map}} = \sum_{i=1}^M \mathbbm{1}_\text{map}(i) R_{\text{map},i}.
\end{equation}

\paragraph{Speed.}
We employ a speed limit guide function to ensure that agents adhere to both a predefined maximum speed $v_\text{max}$ and a minimum speed $v_\text{min}$, promoting safe and controlled behavior.
Let $v_i^t$ denote the speed for agent $i$ at timestamp $t$, and the speed limit guide is defined as the amount by which the predicted speed deviates from the allowable range $[v_\text{min}, v_\text{max}]$, computed as: 
\begin{align} 
    R_{\text{speed},i}^t  &= \max(v_i^t - v_{\text{max}}, 0) + \max(v_{\text{min}} - v_i^t, 0), \\
    R_{\text{speed},i} &= \displaystyle\sum_{t=1}^T w(t) R_{\text{speed},i}^t, \\
    R_{\text{speed}} &= \sum_{i=1}^M \mathbbm{1}_\text{speed}(i) R_{\text{speed},i},
\end{align} 
where $w(t)$ is an exponential decay weighting function and $\mathbbm{1}_\text{speed}(i)$ is an indicator function that evaluates to $1$ for agents with non-zero velocity and $0$ otherwise.
Finally, the guide $\mathcal{J}$ is defined as follows:
\begin{equation}
    \mathcal{J} = \sum_{i\in\{\text{agent, map, speed}\}} w_i R_i
\end{equation}
In Table~\ref{abl:gen}, the weights are set as $w_\text{agent}=50, w_\text{map}=1$, and $w_\text{speed}=1$ when each respective guide is used.
If a guide is not utilized, its corresponding weight is set to 0.
\subsection{Ego-centric Conversion}\label{sup:ego}
\paragraph{Drawing Map.}
To generate the input maps, we utilize the nuScenes map API to extract relevant map data.
Focusing on an area of $60m \times 60m$ centered around the ego vehicle, we include the following components: [\textit{'drivable area', 'road segment', 'lane', 'ped crossing', 'walkway'}].
We overlay other vehicles onto the map at their corresponding coordinates, accurately rendering each vehicle by incorporating their size and orientation.

\begin{wrapfigure}{r}{0.5\linewidth}
    \centering
    \includegraphics[width=\linewidth]{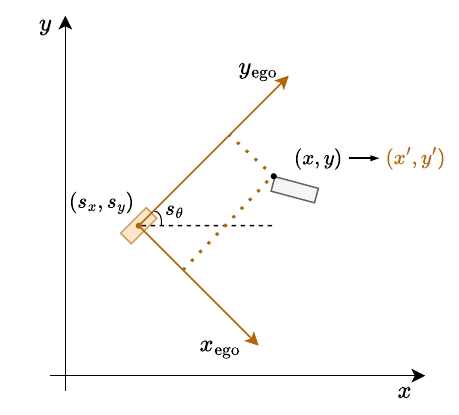}
    \vspace{-2.2em}
    \caption{The coordinate transformation for the ego-vehicle.}
    \label{fig:rot}
\end{wrapfigure}

\paragraph{Rotation Matrix.}
We now describe the rotation matrix used in Equation~\ref{eq:rot}. As illustrated in Figure~\ref{fig:rot}, the new coordinate system for the ego agent is defined by placing the ego's position $(s_x, s_y)$ at the origin and aligning the vehicle's heading direction (north) with the $y$-axis.
First, the position of an arbitrary point $(x,y)$ is translated to $(x-s_x,y-s_y)$ to account for the ego's position. 
Given that the vehicle's heading is rotated counterclockwise by $s_\theta$ radians relative to the original coordinate system, 
\setlength{\intextsep}{0pt}
\setlength{\columnsep}{10pt}
the rotation of the translated point is determined by rotating axis by $\frac{\pi}{2}-s_\theta$ radians clockwise about the origin. This is equivalent to counterclockwise rotation by the same radians.
The standard rotation matrix for rotating a point counterclockwise by an angle $\alpha$ is:

\begin{align} 
        R(\alpha) = \begin{pmatrix}
                      \cos{\alpha} & -\sin{\alpha} \\
                      \sin{\alpha} & \cos{\alpha}
                    \end{pmatrix} 
\end{align}
Substituting $\alpha = \frac{\pi}{2}-s_\theta$ and applying trigonometric identities, the transformation is given by the following equation:
\begin{align} 
        T(x,y;s) = \begin{pmatrix}
                      \sin{s_\theta} & -\cos{s_\theta} \\
                      \cos{s_\theta} & \sin{s_\theta}
                    \end{pmatrix} \times 
                      \begin{pmatrix}
                          x-s_x\\
                          y-s_y
                      \end{pmatrix}.
\end{align}

\section{E2E AD Network Details}\label{sup:arch}
\subsection{Motion Forecasting}
Motion forecasting module predicts motion trajectories $\hat{\mathbf{x}} \in \mathbb{R}^{M \times N \times T \times 2}$ for $M$ agents over $T$ timestamps with possible $N$ series of waypoints $(x,y)$.
We prepare $E_{\text{motion}}$, $E_{\text{agent}}$, and $E_{\text{ego}}$ by encoding normalized anchors and positional information through embedding layers with transformations. 
$E_{\text{motion}}$ contains information from anchors representing general motion patterns (e.g., turning left, going straight); 
$E_{\text{agent}}$ represents motion patterns in each agent's own coordinate system, focusing on motion offsets relative to the agent's current position and orientation; and 
$E_{\text{ego}}$ embeds how each agent's potential trajectory relates to the ego vehicle's position and orientation. 
By feeding these inputs to the MotionEncoder, which consists of Transformers and MLP layers, we generate the motion query embedding $q_\text{motion}$:
\begin{equation} 
q_\text{motion} = \operatorname{MotionEncoder}(E_{\text{motion}}, E_{\text{agent}}, E_{\text{ego}}).
\end{equation}
Using $q_\text{motion}$ and the BEV feature $B$, the MotionDecoder produces the refined motion query $\hat{q}_\text{motion}$,
\begin{equation}
    \hat{q}_\text{motion} = \operatorname{MotionDecoder}(q_\text{motion}, B).
\end{equation}
The final motion prediction $\hat{\mathbf{x}}$ is then computed by feeding $\hat{q}_\text{motion}$ into MLP layers: $\hat{\mathbf{x}} = \operatorname{MLP}(\hat{q}_\text{motion})$.
Simultaneously, the model predicts the probabilities $p_k$ for each trajectory $\hat{\mathbf{x}}_k$ by passing $\hat{q}_\text{motion}$ through MLP layers, followed by a log softmax activation.

\subsection{Occupancy Prediction.}
To estimate the future occupancy of the scene, we predict a sequence of occupancy maps $\hat{O}=\{\hat{O}^1,...,\hat{O}^T\}$, where each $\hat{O}^t \in \mathbb{R}^{H \times W}$ corresponds to the occupancy at timestamp $t$.
First, the instance-level embedding $q_\text{ins} \in \mathbb{R}^{M \times D}$ is derived from $\hat{q}_\text{motion}$ through MLP layers, where $M$ and $D$ are the number of agents and embedding dimension.
At each timestamp $t$, we feed $q_\text{ins}$ into another MLP to generate temporal queries: $q_{\text{temp}}^t = \operatorname{MLP}^{t}(q_{\text{ins}})$.
Simultaneously, the raw BEV feature $B \in \mathbb{R}^{C \times H_\text{bev} \times W_\text{bev}}$ is reduced and downscaled to produce an initial state $B_\text{state}^0 \in \mathbb{R}^{C \times \frac{H_\text{bev}}{4} \times \frac{W_\text{bev}}{4}}$.
A transformer-based decoder, referred to as the OccDecoder, then updates $B_{\text{state}}^{t-1}$ by incorporating $q_{\text{temp}}^{t}$, producing an updated BEV feature $B_{\text{state}}^{t}$ as follows: 
\begin{equation} 
    B_{\text{state}}^{t} = \operatorname{OccDecoder}(B_{\text{state}}^{t-1}, q_{\text{temp}}^{t}).
\end{equation}
After passing through additional upsampling, the final BEV features and the instance queries are fused via an elementwise dot product across the channel dimension, producing the occupancy logits $\hat{O}$.

\section{Training Details}
\label{sup:detail}
\subsection{Map-to-BEV Network}
We first initialize the BEV Query $Q_B$ of Map-to-BEV network from the pre-trained BEVFormer~\cite{li2022bevformer}
We employ ResNet50 as the map encoder, utilizing only the layers up to the point before the pooling layer. 
Transformer encoder comprises $6$ blocks, each including a cross-attention layer, a feedforward network, and normalization layers with residual connections.
Our training setup involves the AdamW optimizer with a cosine annealing scheduler over 20 epochs, including a warm-up phase during the first $5$ epochs. 
We set the learning rate to $5\times10^{-4}$ and apply a weight decay of $0.01$. The momentum parameters are configured with $\beta_1=0.9$ and $\beta_2=0.999$. 

In Table~\ref{abl:arch}, we implement the SwinUNETR architecture following the official implementation.
When utilizing SwinUNETR, we set the feature size to $48$ and the drop rate to $0.2$ and increase the input map size to $800$ for sufficient performance. For other training configurations, we use the same as those used in our experiments with the proposed architecture.

\subsection{Training SynAD}
The E2E AD model consists of perception (tracking, mapping), prediction (motion forecasting, occupancy prediction), and planning modules. 
Existing models include those where each module is serialized~\cite{hu2023planning} and those where they are configured in parallel~\cite{weng2024drive}. 
To appropriately train the E2E AD model using two different types of inputs, multi-camera and map inputs, we suitably combine two design principles.
The perception module detects and tracks each vehicle and pedestrian from the multi-camera input and recognizes the map composition through segmentation. 
In cases where map input is provided instead of multi-camera input, the perception module does not need to operate. 
Therefore, we design the perception module to operate independently by configuring it in parallel.
In the prediction module, motion forecasting uses only BEV features as input, while occupancy prediction uses the BEV features and the output of motion forecasting, $\hat{q}_{\text{motion}}$, as inputs.
The planning module uses only BEV features as input during training, and during inference time, it uses the output of occupancy prediction, $\hat{O}$, along with test-time optimization to reduce the collision rate.

\paragraph{Module Configurations.}
\textit{The motion forecasting module} takes three embeddings($E_{\text{motion}}$, $E_{\text{agent}}$, $E_{\text{ego}}$) along with the BEV feature as inputs. 
The BEV feature has a channel dimension of 256 and spatial dimensions of $200 \times 200$.
$E_{\text{motion}}$, $E_{\text{agent}}$, $E_{\text{ego}}$ each have shapes $M \times N \times 256$, where $M$ is the scenario-dependent number of agents and $N=6$ represents the number of possible paths.
These embeddings are initialized with learnable parameters of size 100, then sliced according to the number of agents in each scenario.
The queries $\hat{q}_{\text{motion}}$ and $q_{\text{motion}}$ share the same shapes as their embeddings.
The module outputs $\hat{\mathbf{x}} \in \mathbb{R}^{M \times N \times T \times 2}$, where $T=12$.
\textit{In the occupancy prediction module}, $\hat{q}_{\text{motion}}$ passes through an MLP layer to form a tensor of shape $M \times 256$, and we also define $q_{\text{temp}}^t \in \mathbb{R}^{M \times 256}$.
\textit{In the planning module}, $q_{\text{plan}}$ and $\hat{q}_{\text{plan}}$ each have shape $1 \times 256$, and $B_a$ retains the same dimensions as $B$.

\paragraph{Loss Configurations.}
When $\lambda_{\text{motion}}$, $\lambda_{\text{occ}}$, and $\lambda_{\text{plan}}$ in Equation~\ref{eq:t_loss} are all set to 1.0, the detailed loss coefficients for each module are as follows:
The loss coefficients $\lambda_{\text{JNLL}}$ and $\lambda_{\text{minFDE}}$ in the motion forecasting are set to 0.5 and 0.25, respectively.
The loss coefficients $\lambda_{\text{dice}}$ and $\lambda_{\text{bce}}$ in the occupancy prediction are set to 1.0 and 5.0, respectively.
In the planning, we used three values for $\delta$: 0, 0.5, and 1.0. 
Accordingly, the loss coefficients $\lambda_{\delta}$ are set to 2.5, 1.0, and 0.25, respectively.

\paragraph{Hyperparameter Configurations.}
We use the AdamW optimizer with a learning rate of $2 \times 10^{-4}$ and a weight decay of 0.01. 
We utilize a cosine annealing learning rate scheduler with a warm-up phase lasting for the initial 500 iterations at a ratio of one-third. 
We employ a cosine annealing learning rate scheduler, performing warm-up for the first 500 iterations at a ratio of one-third, and set the minimum learning rate to $2 \times 10^{-7}$.
The momentum parameters are set to $\beta_1=0.9$ and $\beta_2=0.999$. 

\section{Reporting Rules for Other Works.}
\label{sup:other}
In Table~\ref{exp:main1}, we evaluate our generated paths using the CTG++~\cite{zhong2023language} metrics, and the baseline performances are taken from the same reference.
In Table~\ref{exp:main2}, the reported results for UniAD~\cite{hu2023planning}, VAD, and ParaDrive~\cite{weng2024drive} come from the most recent ParaDrive paper. 
For OccNet~\cite{tong2023scene} and OccWorld~\cite{zheng2025occworld}, we use the performance as reported in OccWorld, selecting the best reported results. 
In Table~\ref{exp:main3}, we obtain the performances of UniAD and VAD via their official codebases, and we rely on our own implementation for ParaDrive due to the lack of publicly available code. 
For the same reason, we adopt the VAD evaluation protocol for planning.

\end{document}